\RequirePackage{fix-cm}
\documentclass[smallextended]{svjour3}       
\smartqed  
\usepackage{graphicx}
\usepackage{epsfig}
\usepackage{amsmath,bm}
\usepackage{algorithm}
\usepackage{algorithmic}
\usepackage{subfigure}
\usepackage{epstopdf}
\usepackage{amssymb}

\usepackage{enumerate}
\usepackage[figuresright]{rotating}

\begin{document}

\title{Cyclic Label Propagation for Graph Semi-supervised Learning
}


\author{Zhao Li         \and
	      Yixin Liu    \and
   	      Zhen Zhang    \and
  	      Shirui Pan    \and
 	      Jianliang Gao    \and
	      Jiajun Bu 
}


\institute{Zhao Li \at
              Alibaba Group \\
              \email{lizhao.lz@alibaba-inc.com}           
           \and
           Yixin Liu \and Shirui Pan \at
              Monash University \\               
           \and
              Zhen Zhang \and Jiajun Bu \at
              Zhejiang University \\           
           \and
              Jianliang Gao \at
              Central South University \\      
}

\date{Received: date / Accepted: date}

\maketitle

\begin{abstract}
Graph neural networks (GNNs) have emerged as effective approaches for graph analysis, especially in the scenario of semi-supervised learning. Despite its success, GNN often suffers from over-smoothing and over-fitting problems, which affects its performance on node classification tasks. We analyze that an alternative method, the label propagation algorithm (LPA),  avoids the aforementioned problems thus it is a promising choice for graph semi-supervised learning. Nevertheless, the intrinsic limitations of LPA on feature exploitation and relation modeling make propagating labels become less effective. To overcome these limitations, we introduce a novel framework for graph semi-supervised learning termed as \underline{\textbf{Cyc}}lic Label \underline{\textbf{Prop}}agation (\textbf{CycProp} for abbreviation), which integrates GNNs into the process of label propagation in a cyclic and mutually reinforcing manner to exploit the advantages of both GNNs and LPA. In particular, our proposed CycProp updates the node embeddings learned by GNN module with the augmented information by label propagation, while fine-tunes the weighted graph of label propagation with the help of node embedding in turn. After the model converges, reliably predicted labels and informative node embeddings are obtained with the LPA and GNN modules respectively. Extensive experiments on various real-world datasets are conducted, and the experimental results empirically demonstrate that the proposed CycProp model can achieve relatively significant gains over the state-of-the-art methods. 
\end{abstract}

\section{Introduction}\label{sec:intro}
Graph-structured data is pervasive in various applications, ranging from citation networks, social networks to E-commerce networks. Mining knowledge in graphs such as predicting node properties is desirable and meaningful to both academic and industrial communities. For example, given an academic citation network, we may be interested in predicting the research area of an author. Making such predictions has become the focus of graph analysis which broadly includes graph classification \cite{kudo2005application}, link prediction \cite{liben2007link} and community detection \cite{leskovec2010empirical}, {\it etc}. Among various graph analysis problems, semi-supervised node classification for graphs is an essential and widespread task, and it has attracted great interests \cite{Yang2016Revisiting,kipf2017semi,velivckovic2018graph}.

Graph representation learning is an effective technique for tackling this task. Early shallow approaches \cite{perozzi2014deepwalk,grover2016node2vec} typically follow a two-step framework, which aims to learn a continuous, compact, and low-dimensional embedding (vector) for each node in the graph. These embeddings are further fed into a classifier to infer the labels of nodes. Since the node representations are not optimized for the specific classifier, this two-step process will inevitably lead to sub-optimal performance. More recently, several semi-supervised graph neural networks (GNNs) \cite{kipf2017semi,velivckovic2018graph,wu2019simplifying} were proposed. They utilize deep learning techniques \cite{lecun2015deep} such as convolution or attention mechanism to encode both the local graph structure and node attributes to generate embeddings, which are then followed by a prediction layer (e.g., softmax or logistic sigmoid function) for the classification purpose. Due to the powerful feature extraction ability of deep learning and the integrated end-to-end framework, they have achieved state-of-the-art performance in the node classification task. 

While these GNNs approaches have become the de facto solution for graph semi-supervised learning, they still suffer from two shortcomings, the \textit{over-smoothing} and \textit{over-fitting} problems, due to the inherent training and test procedure for semi-supervised learning \cite{rong2019dropedge}. First, GNN essentially employs a message passing neural network with neighborhood representation aggregation to train a model and map the feature space into the label space \cite{gilmer2017neural,ying2019gnnexplainer}. When the network architecture goes deep, due to the excessive aggregation, all nodes’ representations tend to converge to a stationary point, resulting in the indistinguishable representations of nodes in different classes \cite{chen2020measuring}. Such an issue is denoted as \textit{over-smoothing} problem which seriously affects the performance of GNNs \cite{li2018deeper}. Second, GNN aims to minimize a loss function over the labeled nodes, which are typically very limited in semi-supervised learning. Therefore, GNNs easily \textit{overfit} the training labels, leading to a degenerated generalization performance.  

In semi-supervised learning, one alternative and promising learning paradigm is the label propagation algorithm (LPA) \cite{zhu2003semi,zhou2004learning}. Different from representation learning or GNNs, LPA builds up a graph over the labeled and unlabeled data, where edges in the graph connect semantically similar nodes with its weights reflecting how strong the similarities are. Then, LPA infers the labels of unlabeled nodes by propagating known labels through neighbors of each node.  The edge weights in LPA are often set heuristically based on the observed node attributes (e.g., using Gaussian kernel function).

LPA has nice properties that can avoid both the \textit{over-smoothing} and \textit{over-fitting} problem faced by GNNs. Concretely, GNNs learn the feature mapping function with multiple aggregations which cause \textit{over-smoothing}, but LPA directly spreads the labels on a graph without involving the feature learning process. As a result, LPA would not lead to indistinguishable node representations. Furthermore, LPA classifies nodes by propagating the labels instead of training a classifier to fit the limited training data, thus they can prevent from \textit{over-fitting}.

However, propagating labels effectively is not trivial, since the classic LPA still has the following intrinsic limitations. 
\begin{itemize}
	\item \textit{Limitation 1: Limited capacity to exploit features.} Classic LPA  derives edge weights from original high dimensional node attribute space where a large portion of sparse, redundant or noisy information is contained. These approaches cannot exploit the expressiveness of features effectively. Furthermore, computing the similarity of the raw attribute directly may lead to noisy weight values and the loss of key information. 
	\item \textit{Limitation 2: Hardly capture the strength of relation corresponding to the labels.} The edge weight is computed as a separate step from the label propagation in LPA. As a result, the label information is ignored in capturing the strength of the relation. As the edge weights are only calculated once based on the similarity of raw attributes, they cannot be updated reversely by the label propagation process. The fixed edge weights limit the classification performance for semi-supervised classification.
\end{itemize} 

\begin{figure}
	\includegraphics[width=0.9\textwidth]{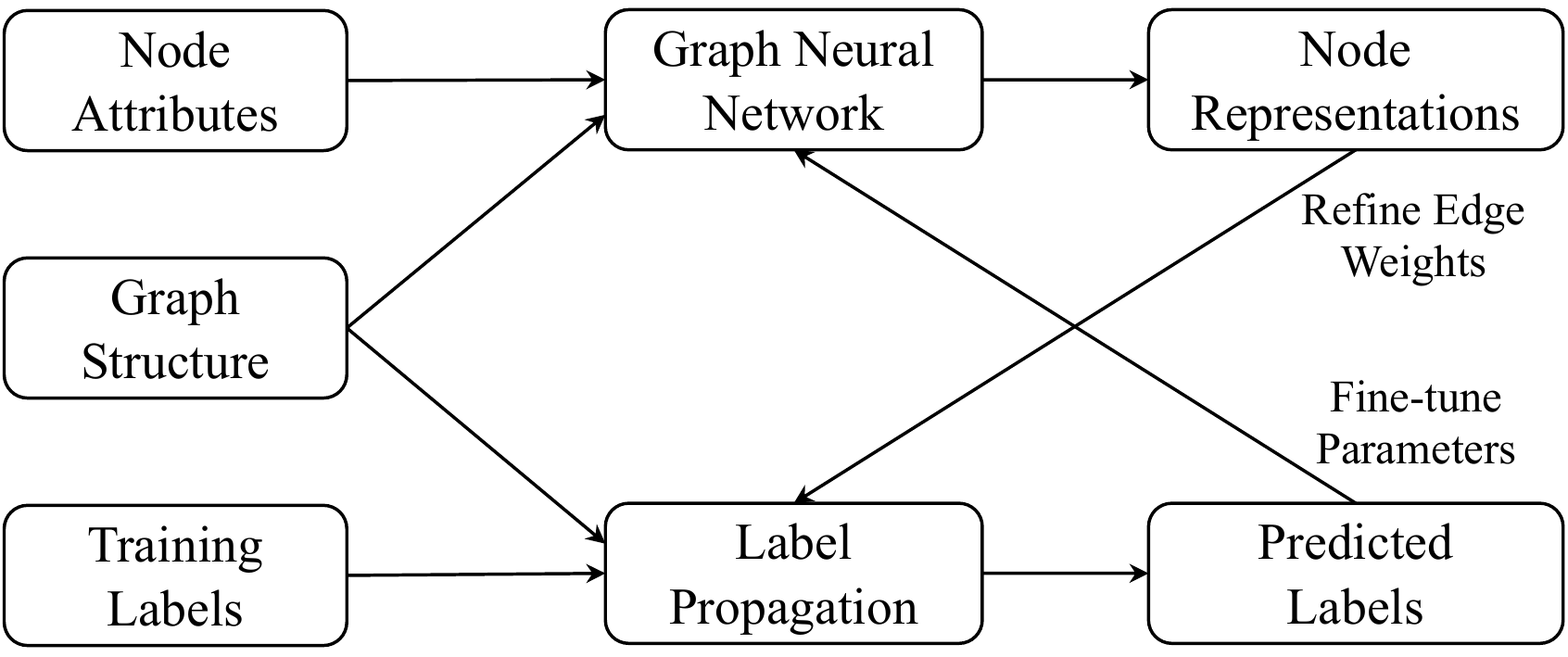}
	\caption{A concept map of our proposed framework. The node representations and the predicted labels update each other in a cyclic manner.}
	\label{fig:simp_model}       
\end{figure}

To overcome the aforementioned limitations, in this paper, we propose a novel framework for graph semi-supervised learning named \underline{\textbf{Cyc}}lic Label \underline{\textbf{Prop}}agation (\textbf{CycProp} for abbreviation).  Our theme is to integrate GNNs into the process of label propagation in a cyclic and mutually reinforcing manner to exploit the advantages of both GNNs and LPA algorithms. More specifically, to overcome \textbf{Limitation 1}, we derive a novel label-adaptive graph neural network module to learn low-dimensional embeddings of nodes in a graph. To enhance the representation power of the embedding, we exploit the highly reliable labels obtained from label propagation in the negative sampling process, so that the label information can be nicely injected into the node embedding component. For \textbf{Limitation 2}, we develop an embedding-adaptive label propagation module, which utilizes the node embeddings to refine the edge weights for label propagation. With the label information injected in the node embeddings, the weights essentially capture the strength of the relations corresponding to the labels. A self-paced learning manner is devised to adaptively control the cyclic learning process, where the embedding learning and edge refining are optimized alternatively to mutually benefit each other in our framework. Once the model has converged, the unknown node labels can be obtained on-the-fly without training an extra classifier or performing a sophisticated inference procedure. A concept map of our framework is given in Figure \ref{fig:simp_model}.

To summarize, the main contributions are as follows:

\begin{itemize}
	\item We propose CycProp, a unified graph semi-supervised learning framework which exploits the advantages of both GNN and LPA. By updating node representation and weighted graph in a cyclic and mutually reinforcing way, our proposed framework can obtain label estimations and node embeddings simultaneously. 
	\item We design a novel label-adaptive graph neural network module for graph representation learning, which leverages not only structure context but also self-adaptive augmented label context to learn the node embeddings.
	\item We conduct extensive experiments on various datasets to demonstrate the effectiveness of CycProp and its superiority compared to a range of state-of-the-art methods.
\end{itemize}

\section{Related Work}\label{sec:related}
\subsection{Graph Representation Learning}
Graph embedding, an important branch of graph representation learning aims to embed nodes into latent vector spaces, where the inherent properties of the graph are preserved. Motivated by the success of Word2vec \cite{Mikolov2013befficient,mikolov2013distributed}, Skip-gram model is employed and adapted from word embeddings to node embeddings based on the graph topology. For instance, DeepWalk \cite{perozzi2014deepwalk} and node2vec \cite{grover2016node2vec} use different sampling strategies to generate random walk sequences, which are then fed into Skip-gram model to learn low-dimensional embedding vectors. LINE \cite{tang2015line} optimizes both the first-order and second-order proximity preserving objectives. While the above methods only utilize the graph structure information, some recent approaches attempts to consider preserving both the structure and attribute proximities in a unified space. For example, SNE \cite{Liao2017Attributed} leverages a deep neural network architecture to capture the complex interrelations between graph structure and node attributes information. GraphSAGE \cite{hamilton2017inductive} generates embeddings by sampling and aggregating attributes from nodes' local neighborhoods in an inductive setting. EP \cite{duran2017learning} tries to learn vector representations by utilizing a propagation design. Another category of graph embedding algorithms follows a semi-supervised manner, in which available information includes not only node attributes but also node labels. Among them, TriDNR \cite{Pan2016Tri} models tri-party information sources including node structure, node attributes and node labels to jointly learn node representations. Planetoid \cite{Yang2016Revisiting} simultaneously optimizes the prediction of known labels and its corresponding graph contexts to learn node's representations. 

Recently, graph neural networks (GNNs) have raised extensive attention and attained state-of-the-art performance in several graph analysis applications, especially in semi-supervised node classification task. By applying deep learning techniques \cite{lecun2015deep} to non-Euclidean domains, GNNs can learn node representation from high-dimensional feature space and predicted labels simultaneously in an end-to-end way \cite{wu2020comprehensive}. GCN \cite{kipf2017semi}, a representative model of GNNs, performs spectral convolutions on graph to encode both local graph structure and attributes of nodes into hidden representations. GAT \cite{velivckovic2018graph} applies attention mechanism to attend over node’s neighborhood contents for generating node embeddings. SGC \cite{wu2019simplifying} simplifies GCN by removing the non-linearity and also achieves competitive performance. Unfortunately, GNNs for node classification usually suffer from two main obstacles, \textit{over-fitting} and \textit{over-smoothing}, which seriously hurt the performance of models \cite{rong2019dropedge}.

\subsection{Label Propagation Algorithm}
Label propagation algorithm (LPA) has been proposed as an efficient method to learn missing labels for graph data in a semi-supervised setting. GFHF \cite{zhu2003semi} learns the predicted labels by optimizing the harmonic functions based on a Gaussian random field model. LLGC \cite{zhou2004learning} considers local and global prior consistency through combining a smoothness constraint and a fitting constraint. LNP \cite{wang2006label} studies the graph construction by approximating the whole graph with linear neighborhood structures, where labels are propagated to the remaining unlabeled nodes using the constructed graph. DLP \cite{wang2013dynamic} deals with multi-label propagation problem via considering the label correlation information. Moreover, inspired by LPA's formulation, other common approaches try to train a supervised learner to classify data features while regularizing it using graph information. For example, manifold regularization \cite{belkin2006manifold} trains a support vector machine with a graph Laplacian regularizer. LSHM \cite{jacob2014learning} addresses the node classification task in heterogeneous social networks via the learned node representations. 

Several recent works also exploit LPA to neural networks. For instance, NGM \cite{bui2018neural} utilizes the power of neural networks and constrains neighborhood nodes to learn similar representations for classification. LP-DSSL \cite{Iscen2019Label} utilizes label propagation to generate pseudo-labels for the unlabeled data, which expands the training sample set for neural network training. GCN-LPA \cite{wang2020unifying} builds a GCN with learnable edge weights and views LPA as regularization to assist the GCN in learning proper edge weights. Our proposed CycProp also combines label propagation and neural network, but has several essential differences with the above methods: (1) All of the aforementioned methods predict the unknown label by neural networks, whereas CycProp makes classification prediction by label propagation, which effectively avoids \textit{over-smoothing} and \textit{over-fitting} problems; (2) In the above methods, the main component of objective functions is the cross-entropy loss function and LPA serves as a regularization term or pseudo-label generator, while we set the label propagation loss as the main objective and also design a structure-label-aware graph embedding loss function.

\section{Cyclic Label Propagation for Graph Semi-supervised Learning}\label{sec:model}
In this section, we first define the notations and present our problem formulation. Then, we introduce the two major components in our unified framework: (1) label-adaptive graph neural network module, and (2) embedding-adaptive label propagation module. After that, a joint training framework that integrates the two components is presented.

\subsection{Notations and Problem Formulation}
Given an attributed graph $\mathcal{G} = (\mathcal{V}, \mathcal{E}, \mathbf{X})$, where $\mathcal{V} = \{v_1, \cdots, v_l, $ $v_{l+1}, \cdots, v_n\}$ and $\mathcal{E}$ denote the set of nodes and edges, respectively; $\mathbf{X} \in \mathbb{R}^{n \times m}$ is a matrix that represents all node attributes, and $\mathbf{x}_i \in \mathbb{R}^{m}$ denotes the attributes affiliated with node $v_i$. Let label set $\{1, 2, \cdots, K\}$ represent different classes of labels and $\mathbf{Y} \in \mathbb{R}^{l \times K}$ be a label matrix, in which $\mathbf{y}_i \in \mathbb{R}^{K}$ denotes the label distribution of node $v_i$, i.e., if $v_i$ belongs to class $j$, then $y_{ij} = 1$, otherwise $y_{ij} = 0$. The first $l$ nodes $v_i \  (1 \le i \le l)$ are labeled, and the remaining nodes $v_u \  (l+1 \le u \le n)$ are unlabeled. With the above notations, we formally define our problem as follows.
\begin{definition}
	Given an attributed graph $\mathcal{G}$, partially labeled nodes $\{v_1, \dots, v_l\}$, and the desired node embedding dimension $d$. Our goal is to learn the label assignments $\mathbf{F} \in \mathbb{R}^{n \times K}$ and node embeddings $\mathbf{E} \in \mathbb{R}^{n \times d}$ simultaneously. Each node has a probability distribution over the set of labels.
\end{definition}

\subsection{Label-adaptive Graph Neural Network Module}\label{sec:label-adaptive}
To learn meaningful node embeddings, it's desirable to incorporate various available graph information. Different from \cite{tang2015pte} that models the attribute information as augmented nodes or \cite{huang2017accelerated} that use node attributes to calculate a similarity matrix, we designed a label-adaptive graph neural network module to capture node's deep semantics.

Specifically, we generate node embeddings as follows,
\begin{equation}
\mathbf{e}_i = g_{\bm{\theta}}(\mathbf{x}_i),
\end{equation}
where $g_{\bm{\theta}}(\cdot)$ can be any kind of GNNs \cite{kipf2017semi,velivckovic2018graph,wu2019simplifying,wu2020comprehensive}, and $\bm{\theta}$ is the parameter set. These methods typically work by propagating representations throughout the graph. Here, we choose GraphSAGE \cite{hamilton2017inductive} as our graph neural network module in the experiments, due to its effectiveness and efficiency. Then, we optimize it in a label-adaptive manner by predicting its associated graph context.  Formally, let $(i, c)$ represent the node-context pair, i.e., node $v_c$ is the graph context of node $v_i$, and our goal is to minimize the following log softmax probability,

\begin{equation}
-{\rm log} \ \sigma(\mathbf{e}_c^{\rm T}\mathbf{e}_i) - \sum_{s=1}^{s_{neg}}\mathbb{E}_{v_j \sim P_n(v)}[{\rm log} \ \sigma(-\mathbf{e}_j^{\rm T}\mathbf{e}_i)],
\end{equation}

where $P_n(v) \propto d_v^{3/4}$ as suggested in \cite{mikolov2013distributed}, and $d_v$ is the degree of node $v$. Then, the goal is indeed transformed to classify the node-context pairs $(i,c)$ into positive context $(\gamma = +1)$ or negative context $(\gamma = -1)$ sampled from a noisy distribution. Therefore, the graph embedding loss with negative sampling can be rewritten as,

\begin{equation}
\label{ge_loss}
\mathcal{L}_{GE} = - \sum_{i=1}^n\mathbb{E}_{(i,c,\gamma)}{\rm log}\ \sigma(\gamma\mathbf{e}_{c}^{\rm T}\mathbf{e}_i),
\end{equation}
where $\sigma(\cdot)$ is the sigmoid function, i.e., $\sigma(x) = 1/(1+e^{-x})$.

\begin{algorithm}
	\caption{Structure-label-aware Graph Context Sampling}
	\label{context_sampling_algo}
	\begin{algorithmic}[1]
		\REQUIRE {Graph $\mathcal{G}$, indicator vector $\bm{\varphi}$, negative sample number $s_{\rm neg}$, probability $r$, noise distribution $P_n(v)$}.
		\ENSURE {Graph context pairs $(i,c, \gamma)$}.
		\STATE Draw a number $rnd \sim Uniform(0, 1)$.
		\IF {$rnd < r$} 
		\STATE $//$ {\it sampling structure context}
		\STATE Uniformly sample a positive pair $(i,c,+1)$ with $v_c$ from node $v_i$'s neighborhood.
		\STATE Sample $s_{\rm neg}$ negative pairs $(i,c,-1)$ with $c$ from $\mathcal{V}$ corresponding to $P_n(c)$.
		\ELSE
		\STATE $//$ {\it sampling label context}
		\STATE Uniformly sample a node $v_i$ from nonzero elements in $\bm{\varphi}$.
		\STATE Uniformly sample a positive pair $(i,c,+1)$ with $\varphi_c = 1$ and $y_i = y_c$.
		\STATE Sample $s_{\rm neg}$ negative pair $(i,c,-1)$ with $c$ from $\mathcal{V}$  with $\varphi_c = 1$ and $y_i \ne y_c$ corresponding to $P_n(c)$.
		\ENDIF
	\end{algorithmic}
\end{algorithm}

We now present how to generate $(i, c, \gamma)$ graph context pairs with a structure-label-aware sampling process. We develop two graph context sampling mechanisms, which is depicted in Algorithm \ref{context_sampling_algo}.

\vspace{1mm}
\noindent\textbf{Structure aware graph context sampling}: The first type is based on the graph structure, which encodes the structure information and regards the connected nodes as positive node-context pairs. 

\vspace{1mm}
\noindent\textbf{Label aware graph context sampling}: The second type is based on the label set, which injects label information into the context and treats the nodes having the same labels as positive node-context pairs. Moreover, an indicator variable of $\bm{\varphi}$ is introduced to control the label related graph context. We first initialize the $0/1$ indicator vector $\bm{\varphi} \in \mathbb{R}^n$ with the known label set, i.e., $\varphi_i = 1$ means node $v_i$ is a label context candidate. Then we augment it through the label propagation module, where the learned highly reliable labeled nodes will be gradually incorporated into $\bm{\varphi}$ to expand the label context candidates. The generated context pairs will be dynamically refreshed during the training process, due to the updating of parameter $\bm{\varphi}$ and node labels. The details about how to model the indicator vector $\bm{\varphi}$ will be described in the following subsection.

\subsection{Embedding-adaptive Label Propagation Module}\label{sec:embedding-adaptive}
In order to overcome the limitations of existing label propagation approaches \cite{zhou2004learning,wang2006label,wang2013dynamic}, we propose to infer the edge weights in an informative embedding space via a mutually reinforcing manner. In detail, for each node $v_i$, its corresponding embedding vector $\mathbf{e}_i \in \mathbb{R}^{d}$ is obtained from the label-adaptive graph neural network module, where $d$ is the embedding dimension ($d \ll |m|$). Basically, edge weights can be calculated based on the following function similarly to \cite{zhu2003semi,zhou2004learning,wang2006label},

\begin{equation}
\label{edge_weights}
s_{ij} = {\rm exp}\left(-\frac{\Vert \mathbf{e}_i -  \mathbf{e}_j \Vert^2}{2\delta^2}\right),
\end{equation}

where $\delta$ is the length scale parameter. With this measure, the estimated edge weights reflect the degree of similarities between each connected node pair, which will be dynamically adjusted in the training procedure as the updating of node embeddings.

The key to semi-supervised learning on graphs is to be in line with the prior consistency, such that the label is smooth over the graph. Following this principle, we devise the regularized objective function for embedding-adaptive label propagation as follows, 

\begin{align}
\label{lp_loss}
\mathcal{L}_{LP} &= \sum_{i=1}^n\sum_{j \in \mathcal{N}(i)}s_{ij}\Vert \mathbf{f}_i - \mathbf{f}_j \Vert_2^2 + \mu \sum_{i=1}^l\Vert \mathbf{f}_i - \mathbf{y}_i \Vert_2^2 \nonumber \\
&+ \sum_{i=1}^n\varphi_i H(\mathbf{f}_i) + \lambda \sum_{i=1}^n-\varphi_i \\
&\text{s.t.} \  f_{ik} \ge 0; \  \sum_{k=1}^Kf_{ik} = 1; \ \varphi_i \in \{0,1\}, \  i = \{1, \cdots, n\}, \nonumber
\end{align}

where $s_{ij}$ is the edge weight between node $v_i$ and $v_j$ calculated according to Equation (\ref{edge_weights}). $\mathbf{y}_i$ indicates the ground truth label and $\mathcal{N}(i)$ represents the neighborhood of node $v_i$. $\mu$ is a trade-off hyper parameter between the smoothness and fitness terms, $\mathbf{f}_i$ is the learned label distribution of node $v_i$. Through this manner, the label propagation procedure benefits from the graph neural network module.

In addition, we introduce a self-paced regularizer \cite{Kumar2010Self} to prioritize label learning task as well as selecting some highly reliable node labels in each training iteration. The regularizer is composed of a Shannon entropy function $H(\cdot)$ and an indicator variable $\bm{\varphi}$. $H(\mathbf{f}_i)$ prevents the uniform label distribution, and is formally defined as follows,

\vspace{-0.05in}
\begin{equation}
H(\mathbf{f}_i) = -\sum_{k=1}^K \ f_{ik}\times{\rm log}(f_{ik}),
\end{equation}

where $f_{ik}$ denotes the probability of node $v_i$ belonging to class $k$. The smaller the Shannon entropy is, the larger the amount of information it contains. Specifically, a small Shannon entropy implies that $\mathbf{f}_i$ has a significantly higher probability value in one specific class. For instance, in an extreme case, if the probability of node $v_i$ in class $k$ is $1$, the Shannon entropy of $\mathbf{f}_i$ will be $0$. The self-paced parameter $\bm{\varphi}$ serves as an indicator vector in graph neural network module, which determines the potential label context based on its current label information. The binary value of $\varphi_i$ indicates whether node $v_i$'s learned label is reliable or not and $\lambda$ acts as a threshold to distinguish the informative labels from the uninformative labels. If the Shannon entropy of $\mathbf{f}_i$ is smaller than the threshold, we set $\varphi_i$ as 1 to indicate that node $v_i$ can be utilized as a label context. As the training process goes on, $\lambda$ is gradually increased such that more learned highly reliable labels can be included in graph embedding procedure to adaptively update node embeddings. In this way, the graph embedding procedure benefits from the label propagation module. Thus, the overall framework naturally forms a closed-loop via a mutually reinforcing manner.

\subsection{CycProp: A Joint Learning Framework}

\begin{figure}
	\centering
	\includegraphics[width=1\textwidth]{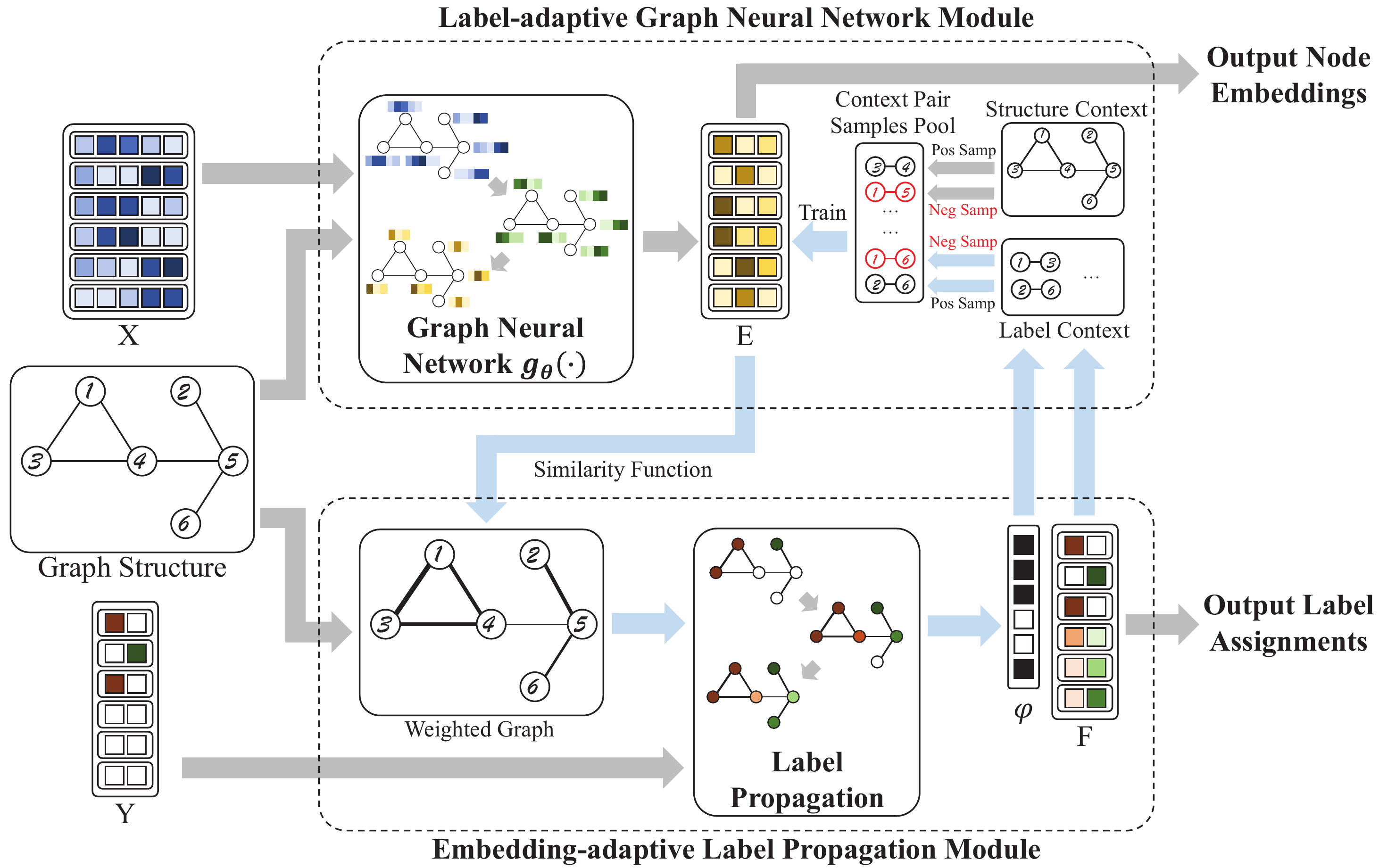}
	\caption{The overall framework of the CycProp model. The nodes 1-3 have known labels while the labels of nodes 4-6 need to be predicted. In the weighted graph, the thicker edges denote larger weights. In the context pair samples pool, the context pairs in black denote the positive samples ($\gamma=1$), and which in red denote the negative samples ($\gamma=-1$).}
	\label{fig:model}
\end{figure}

Figure \ref{fig:model} provides an overview of our proposed CycProp model. The two main components of our proposed framework, label-adaptive graph neural network module and embedding-adaptive label propagation module, output the node embeddings and label assignments (classification results) respectively. Firstly, in the label-adaptive graph neural network, the attribute matrix $\mathbf{X}$ and graph structure information are fed into the graph neural network $g_{\bm{\theta}}(\cdot)$, which generates the node embeddings $\mathbf{E}$. Then, in the embedding-adaptive label propagation module, the weighted graph is computed by a similarity function with the node embeddings $\mathbf{E}$. After that, the predicted label assignments $\mathbf{F}$ and the indicator variable $\bm{\varphi}$ are obtained by label propagation with the weighted graph and known labels $\mathbf{Y}$. Next, the predicted label and indicator are exploited to generate the label context, which is used to train the graph neural network in turn together with the structural context. In this way, the node embeddings $\mathbf{E}$ is updated to generate a more reasonable weight graph, which further optimizes the prediction $\mathbf{F}$ of label propagation. As the blue arrows are shown, by joint learning $\mathbf{E}$ and $\mathbf{F}$ in such a cyclic and mutually beneficial manner, finally, the model can output informative node embeddings as well as accurate label predictions.

The objective function of the proposed model is formulated as the weighted combination of $\mathcal{L}_{LP}$ and $\mathcal{L}_{GE}$ defined in Equations (\ref{ge_loss}) and (\ref{lp_loss}),

\begin{equation}
\label{objective_function}
\mathcal{L} = \mathcal{L}_{LP} + \alpha \mathcal{L}_{GE}.
\end{equation}

Considering the related parameters of both terms in $\mathcal{L}$, the set of learnable parameters in CycProp model is denoted as $\{\mathbf{F},\bm{\varphi},\bm{\theta}\}$. To minimize $\mathcal{L}$, we propose to employ stochastic gradient descent \cite{bottou2010large} and proximal algorithm \cite{parikh2014proximal} to optimize our model via an alternative updating manner. We first give the partial derivative details of some key parameters as follows.

\textbf{Updating F}: We utilize proximal gradient descent \cite{parikh2014proximal,Li2015APG} to solve this constrained optimization problem. In proximal algorithms, the interface to meet the constraint terms is via the proximal operator. Satisfying the non-negative and sum-to-1 constraints in our objective function, i.e., $\mathcal{D} = \{\mathbf{f}|\mathbf{f} \ge 0, \mathbf{f}^{\rm T}\mathbf{1} = 1\}$, is known as computing the projection onto the probability simplex. Here, we employ an efficient algorithm proposed in \cite{wang2013projection} to calculate the proximal operator. The procedure of finding $\mathbf{f} \in \mathcal{D}$ given $\mathbf{z}$ is shown in Algorithm \ref{proj_simplex_algo}.

\begin{algorithm}
	\caption{Projection onto the Probability Simplex}
	\label{proj_simplex_algo}
	\begin{algorithmic}[1]
		\REQUIRE {$\mathbf{z} \in \mathbb{R}^K$}.
		\STATE Sort $\mathbf{z}$ into $\mathbf{u}$: $u_1 \ge u_2 \ge \cdots \ge u_K$.
		\STATE Find $\rho = \max \{1 \le j \le K: u_j + \frac{1}{j}(1 - \sum_{i=1}^ju_i) > 0\}$.
		\STATE Define $\eta = \frac{1}{\rho}(1 - \sum_{i=1}^{\rho}u_i)$.
		\ENSURE {$\mathbf{f}$, s.t. $f_i = \max \{z_i + \eta, 0\}$, $i = 1,2,\cdots,K$}.
	\end{algorithmic}
\end{algorithm}
Then we have the following equation as our proximal operator,
\begin{equation}
\label{proximal_f}
\mathbf{f} = \mathbf{{\rm \mathbf{prox}}}_{\mathcal{D}}(\mathbf{z}) = (\mathbf{z} + \eta\mathbf{1})_{+},
\end{equation}
where $(x)_{+} =\max \{x, 0\}$ and $\eta$ is computed according to the procedure shown in Algorithm \ref{proj_simplex_algo}. Note that the proximal operator always keeps our updated label distribution satisfied with the constraint $\mathcal{D}$. The partial derivative of $\mathbf{f}_i$ can be formulated as,
\begin{equation}
\frac{\partial \mathcal{L}}{\mathbf{f}_i} = \sum_{j \in \mathcal{N}(i)}2s_{ij}(\mathbf{f}_i - \mathbf{f}_j) + 2\mu(\mathbf{f}_i - \mathbf{y}_i)\cdot\mathbb{I}(i) - \varphi_i({\rm log}(\mathbf{f}_i) + \mathbf{1}),
\end{equation}
where $\mathbb{I}(i)$ is an indicator function to indicate whether $i$ is a labeled node or not and $\mathbf{1} \in \mathbb{R}^K$ represents the all one vector. Based on the proximal gradient method, parameter $\mathbf{f}_i$ can be updated as following,
\begin{equation}
\label{label_param_update}
\mathbf{f}_i = \mathbf{{\rm \mathbf{prox}}}_{\mathcal{D}}(\mathbf{f}_i - lr \cdot\frac{\partial \mathcal{L}}{\mathbf{f}_i}),
\end{equation}
where $lr$ denotes the learning rate in gradient descent.

\textbf{Updating} $\bm{\varphi}$: We first relax $\bm{\varphi}$ to take any real value in the interval $[0,1]$. Then the partial derivative of $\varphi_i$ can be derived as,
\begin{equation}
\label{self-paced}
\frac{\partial{\mathcal{L}}}{\partial{\varphi_i}} = H(\mathbf{f}_i) - \lambda.
\end{equation}
Since the optimal value of $\bm{\varphi}$ is constrained to either 1 or 0 for all samples, the closed-form solution to update $\varphi_i$ is,
\begin{eqnarray}
\label{self-paced-solution}
\varphi_i = \begin{cases} 1, &H(\mathbf{f}_i) \le \lambda\\
0, & \text{Otherwise} \end{cases}.
\end{eqnarray}

\begin{algorithm}
	\caption{Joint Training Framework}
	\label{joint_learning_algo}	
	\begin{algorithmic}[1]	
		\REQUIRE {Graph $\mathcal{G} = (\mathcal{V}, \mathcal{E}, \mathbf{X})$, labels $\mathbf{Y}$, batch size $B$, iteration numbers $T_1, T_2$, self-paced hyper parameter $\lambda$.}	
		\ENSURE { \ \newline
			(1) Optimal node embeddings $\mathbf{E} \in \mathbb{R}^{n \times d}$; \newline
			(2) Predicted probability distributions $\mathbf{F} \in \mathbb{R}^{n \times K}$.}	
		\STATE Randomly initialize the parameters $\mathbf{\theta}$ of GNN.
		\STATE Initialize $\mathbf{F}$ by concatenating $\mathbf{Y}$ and zero matrix $\mathbf{0}^{(n-l) \times K}$.
		\STATE Initialize $\bm{\varphi}$ as indicator vector with labeled nodes.
		\WHILE {not converge}
		
		\FOR{$t = 1,2, \cdots, T_1$}
		\STATE Sample $B$ graph context pairs by Algorithm \ref{context_sampling_algo} with indicator vector $\bm{\varphi}$.
		\STATE Update the parameters by taking a gradient step based on Equation (\ref{ge_loss}).
		\ENDFOR
		
		\STATE Obtain $\mathbf{E}$ and compute $\mathbf{S}$ based on Equation (\ref{edge_weights}) to fine-tune edge weights.
		
		\FOR{$t = 1,2, \cdots, T_2$}
		\STATE Update learned probability distribution $\mathbf{F}$ by taking a gradient step based on Equation (\ref{label_param_update}).
		\ENDFOR
		
		\STATE Update $\bm{\varphi}$ based on Equation (\ref{self-paced}) and (\ref{self-paced-solution}).
		\STATE Augment $\lambda$.
		\ENDWHILE
		
	\end{algorithmic}
	
\end{algorithm}

Note that, calculating the partial derivative of parameter set $\bm{\theta}$ in graph neural network module is quite easy, thus we omit the detailed mathematical derivations here due to space limitation. After we have obtained the derivatives of all the parameters, the whole optimization procedure can be efficiently performed via back-propagation. To sum up, the procedure of the joint training framework for CycProp model is depicted in Algorithm \ref{joint_learning_algo}. First, the parameters $\mathbf{F},\bm{\varphi},\bm{\theta}$ are initialed. Then, in each iteration step, the graph neural network module and the label propagation module are updated for $T_1$ steps and $T_2$ steps respectively. In the end of an iteration step, the indicator $\bm{\varphi}$ is updated and the self-paced hyper parameter $\lambda$ is increased. When the algorithm converges, the predicted classification results $\mathbf{F}$ and node embeddings $\mathbf{E}$ are returned simultaneously.

\section{Experiments}\label{sec:experiments}
In this section, we report the results of our experiments to verify the effectiveness of our proposed CycProp model. We first describe the datasets and experimental setups in detail, and then we present the results with insights.

\subsection{Datasets}
We adopt three citation networks and two social networks for empirical studies. Statistics of the five datasets are summarized in Table \ref{datasets} with more descriptions as follows,

\begin{table}[h]
	\centering
	\begin{tabular}{ c|c|c|c|c }
		\hline
		Datasets  & $\#|\mathcal{V}|$ & $\#|\mathcal{E}|$  & $\#|Attrs|$ & $\#|{L}|$\\
		\hline
		Cora            &   2,485   & 5,069     & 1,433    	& 7 \\
		Citeseer        &   2,110   & 3,719     & 3,703   	& 6 \\
		Pubmed			&	19,717	& 44,338	& 500		& 3 \\
		Blogcatalog     &   5,196   & 171,743   & 8,189    	& 6 \\
		Flickr			&	7,575	& 239,738	& 12,047	& 9 \\
		\hline
	\end{tabular}
	\caption{Statistics of the datasets.}
	\label{datasets}
\end{table}

\begin{itemize}
	\item \textbf{Citation Networks}. Cora, Citeseer and Pubmed\footnote{http://linqs.cs.umd.edu/projects/projects/lbc} \cite{lu2003link} are three available public datasets, which are composed of scientific publications. In these networks, nodes represent published papers and edges denote citation relationships. Node labels indicate the categories to which each paper belongs and the text contents are treated as node attributes. We remove papers which have no connection in the network and extract the maximum connected component.
	\item \textbf{Social Networks}. Blogcatalog and Flickr\footnote{http://socialcomputing.asu.edu/pages/datasets} \cite{tang2009relational} are two typical social networks, where nodes represent users and links denote the following relationships. In social networks, users usually generate personalized contents such as posting blogs or sharing photos with tag descriptions, thus these text contents are regarded as node attributes. We set the groups that users joined as labels, and users with no follower or predefined category have been removed.
\end{itemize}

\subsection{Competitors}
We compare the proposed CycProp model against several state-of-the-art baselines that can be categorized into the following groups:
\begin{itemize}
	\item \textbf{Classical LPA.} These methods perform label propagation based on the edge weights calculated from the original attribute vectors. We consider three popular methods GFHF \cite{zhu2003semi}, LLGC \cite{zhou2004learning} and DLP \cite{wang2013dynamic} as our compared algorithms.
	
	\item \textbf{Unsupervised Node Representation Learning.} Methods of this group first employ graph embedding techniques to learn the optimal node representations and then classify each node independently in the latent representation space. These approaches can be further classified into the following two classes:
	
	\textbf{1) Structure-only.} In this group, we choose three baselines DeepWalk \cite{perozzi2014deepwalk}, LINE \cite{tang2015line} and node2vec \cite{grover2016node2vec}, which utilize graph topological information only, and the node attributes are not taken into consideration.
	
	\textbf{2) Attribute + Structure.} This category of algorithms aims to encode both node attributes proximity and graph structure proximity into the latent representation space. We consider three recently proposed methods SNE \cite{Liao2017Attributed}, EP \cite{duran2017learning} and GraphSAGE \cite{hamilton2017inductive} as our baselines.
	
	\item \textbf{Semi-supervised Node Representation Learning.} Methods in this group further leverage additional label information to model the underlying representations. These approaches can be further classified into the following three classes:
	
	\textbf{1) Semi-supervised Node Embedding.} This kind of methods train embeddings for each nodes with the supervision of label data. Planetoid \cite{Yang2016Revisiting} is one of the typical methods and is selected as the compared method.    
	
	\textbf{2) GNN.} By employing deep learning techniques\cite{lecun2015deep}, GNNs learn node representation as well as classifier in a joint and end-to-end way. We select two representative GNN models GCN \cite{kipf2017semi} and GAT \cite{velivckovic2018graph} as our baselines.   
	
	\textbf{3) Neural Network with Label Propagation.} This group of methods combine neural network with label propagation to enhance the classification performance.  We choose three recently proposed methods NGM \cite{bui2018neural}, GCN-LPA \cite{wang2020unifying} and LP-DSSL \cite{Iscen2019Label} as our competitors. 
\end{itemize}

For baseline algorithms, we use the source code released by the authors and the dimension of node embedding is set as $64$ for all methods in all datasets. Specially, LP-DSSL \cite{Iscen2019Label} is originally designed with convolutional neural network for image classification. To adapt it to the network datasets, we replace the convolutional neural network with a two-layer GCN \cite{kipf2017semi}, and use the topology graph in dataset for label propagation instead of KNN graph. We randomly sample $30\%$ labeled nodes as training set, and another $100$ labeled nodes are sampled as a validation set to tune the hyper parameters. The remaining unlabeled nodes are used to test the performance of different algorithms. For our proposed CycProp model, the graph neural network's structure used in our experiments is a 2-hops neighborhood aggregation with dimension of 128 and 64, respectively. The sampled neighborhood size and negative context size are set as 10 for all datasets. We use rectified linear units as the activation function to introduce the non-linearity. To measure the classification result, we utilize both Micro-F1 (Mi-F1) and Macro-F1 (Ma-F1) as evaluation metrics. For unsupervised node representation methods, the learned node representations are regarded as features to train a one-vs-rest logistic regression classifier implemented by scikit-learn\footnote{https://scikit-learn.org/stable/}. Evaluations are conducted by repeating $10$ times with resampled labels, then the average score and its standard derivation are recorded as the final result.

\subsection{Results and Analysis}

\begin{sidewaystable}
	\vspace{3in}
	\centering
	\begin{tabular}{ c|c c|c c|c c|c c|c c }
		\hline
		Datasets & \multicolumn{2}{c|}{Cora}  & \multicolumn{2}{c|}{Citeseer} & \multicolumn{2}{c|}{Pubmed} & \multicolumn{2}{c|}{Blogcatalog} & \multicolumn{2}{c}{Flickr} \\
		\hline
		Metrics(\%)& Mi-F1 & Ma-F1 & Mi-F1 & Ma-F1  & Mi-F1 & Ma-F1 & Mi-F1 & Ma-F1 & Mi-F1 & Ma-F1 \\
		\hline
		LLGC   & 77.6(0.7) & 76.5(0.7) & 62.9(0.8)  & 58.2(0.6) & 82.6(0.3)  & 81.4(0.3) & 65.8(0.4) & 64.9(0.3) & 54.3(0.3) & 53.9(0.3) \\
		GFHF& 81.4(0.8) & 80.7(0.5) & 65.1(0.6) & 61.4(0.5)  & 83.5(0.3) & 82.7(0.2) & 68.4(0.5) & 67.7(0.4) & 58.1(0.3) & 57.7(0.2) \\
		DLP& 78.9(0.5) & 77.6(0.4) & 63.8(0.5) & 59.7(0.4)  & 83.1(0.4) & 82.4(0.3) & 67.2(0.4) & 66.6(0.4) & 57.5(0.4) & 57.1(0.3) \\
		\hline
		DeepWalk& 75.8(0.5) & 74.8(0.4) & 57.3(0.6)  & 53.9(0.5) & 81.6(0.2) & 80.4(0.2) & 64.5(0.5) & 63.9(0.4) & 48.7(0.4) & 48.2(0.4)  \\
		LINE& 73.6(0.6) & 71.9(0.5) & 57.1(0.7) & 53.8(0.7) & 81.2(0.4) & 80.0(0.3) & 64.9(0.4) & 64.2(0.3) & 51.4(0.4) & 51.0(0.3) \\
		node2vec  & 74.7(0.6) & 72.5(0.4) & 61.1(0.6) & 56.5(0.4) & 81.1(0.3) & 79.6(0.3) & 65.1(0.4) & 64.5(0.4) & 51.2(0.5) & 49.8(0.4) \\
		\hline
		SNE&  82.1(0.5) & 81.5(0.5) & 67.7(0.8) & 63.3(0.6) & 84.9(0.2) & 83.5(0.2) & 71.6(0.6) & 70.8(0.5) & 61.6(0.4) & 61.1(0.4) \\
		EP& 82.5(1.2) & 81.4(0.8) & 68.5(1.4) & 64.3(1.1)  & 83.8(0.7) & 83.1(0.5) & 73.5(1.3) & 72.7(1.1)  & 63.3(0.9) & 62.9(0.7) \\
		GraphSAGE&  83.6(0.9) & 82.0(0.6) & 69.2(0.8) & 65.7(0.6) & 84.5(0.4) & 83.7(0.3) & 75.8(0.6) & 74.9(0.4) & 66.2(0.5) & 65.8(0.5) \\
		\hline
		Planetoid& 84.2(1.0) & 83.5(0.8) & 73.6(0.5)  & 69.8(0.4)  & 85.5(0.3) & 85.1(0.3) & 78.4(0.3) & 77.9(0.3) & 68.1(0.4) & 67.8(0.4) \\
		\hline
		GCN & 85.5(1.1) & 84.7(0.8)& 77.5(0.8) & 73.5(0.6) & 85.6(0.2) & 85.3(0.2) & 79.2(0.5) & 78.6(0.4) & 69.5(0.2) & 69.0(0.2) \\
		GAT & 86.1(0.8) & 85.3(0.6)& 77.9(0.7) & 73.8(0.6) & 86.4(0.3) & 85.9(0.3) & 80.6(0.4) & 79.8(0.3) & 71.4(0.3) & 71.1(0.3) \\
		\hline
		NGM   & 83.9(0.7) & 82.2(0.5) & 72.9(0.6) & 68.7(0.5) & 85.1(0.3) & 84.7(0.2) & 77.1(0.4) & 76.3(0.3) & 67.7(0.6) & 67.3(0.4) \\
		GCN-LPA   & 86.5(0.6) & 85.2(0.8) & 77.7(0.8) & 72.9(1.3) & 86.1(0.8) & 85.4(0.8) & 79.6(0.7) & 79.4(0.7) & 69.2(0.6) & 68.6(0.6) \\
		LP-DSSL & 87.4(0.6) & 86.2(0.7)& 78.1(1.1) & 73.1(1.1) & 87.5(0.7) & 86.1(0.8) & 79.8(0.6) & 79.1(0.6) & 69.9(0.4) & 69.5(0.5) \\
		\hline
		CycProp& \textbf{87.6}(0.9) & \textbf{86.5}(0.6) & \textbf{79.1}(0.8)  & \textbf{74.6}(0.7) & \textbf{87.7}(0.2) & \textbf{87.1}(0.2) & \textbf{82.3}(0.3) & \textbf{81.6}(0.3) & \textbf{73.6}(0.3) & \textbf{73.2}(0.3)\\
		\hline
	\end{tabular}
	\caption{Classification performance on different datasets. Values in ($\cdot$) are the standard derivation of multiple runs.}
	\label{node_classification}
\end{sidewaystable}

The experimental results of different algorithms over different datasets are presented in Table \ref{node_classification}. To summarize, we have the following observations. 

Generally, we can find that our proposed CycProp beats all baselines in all datasets for all settings. As expected, structure-only node representation learning methods like DeepWalk, LINE and node2vec perform worse than those approaches using node attributes (i.e, SNE, EP and GraphSAGE). The reason is that they only attempt to capture the graph's topology information, which provides very limited information compared to node attributes for node classification task. It's worth noting that these three classical LPAs such as LLGC, GFHF and DLP achieve better performance than structure-only node representation learning baselines. It further indicates the effectiveness of propagating labels on the graph, which gives us a solid foundation for the proposed CycProp model.

In addition, the semi-supervised methods consistently outperform unsupervised baselines with different gains by incorporating partially known node labels into the model. One major reason for the performance lift is because these semi-supervised methods are trained through an end-to-end manner, thus the learned node representations are specifically optimized for the classifier and show powerful discriminability. Finally, our proposed CycProp model is an efficient and direct way to learn the unknown node labels on graph, which aims to propagate labels rather than classifying each node independently. In classical two-step LPAs, edge weights are predetermined and cannot change during the learning process, thus its performance is bounded by the first step. To overcome these limitations, the proposed CycProp integrates GNN and label propagation in a unified framework via a mutually reinforcing manner that results in a great performance boost.

It is remarkable that compared with other methods that combine neural network and label propagation, CycProp shows better performance and generalization. Such a performance gap is caused by two reasons. First, while other methods that make predictions by neural networks, CycProp classifies each node by label propagation, which would not suffer from \textit{over-smoothing} and \textit{over-fitting} problems caused by GNN classifiers. Second, different from other methods that view label propagation as an auxiliary tool for GNNs (such as the regularization term in GCN-LPA or the pseudo-label generator in LP-DSSL), we treat these two algorithms as equal and mutually reinforcing components and integrate them into a joint learning framework. In this way, the advantages of both components are fully leveraged.

\subsection{Ablation Study}
\begin{figure}[!htb]\small
	\vspace{-0.1in}
	\centering
	\subfigure{
		\centering
		\includegraphics[width=0.45\textwidth]{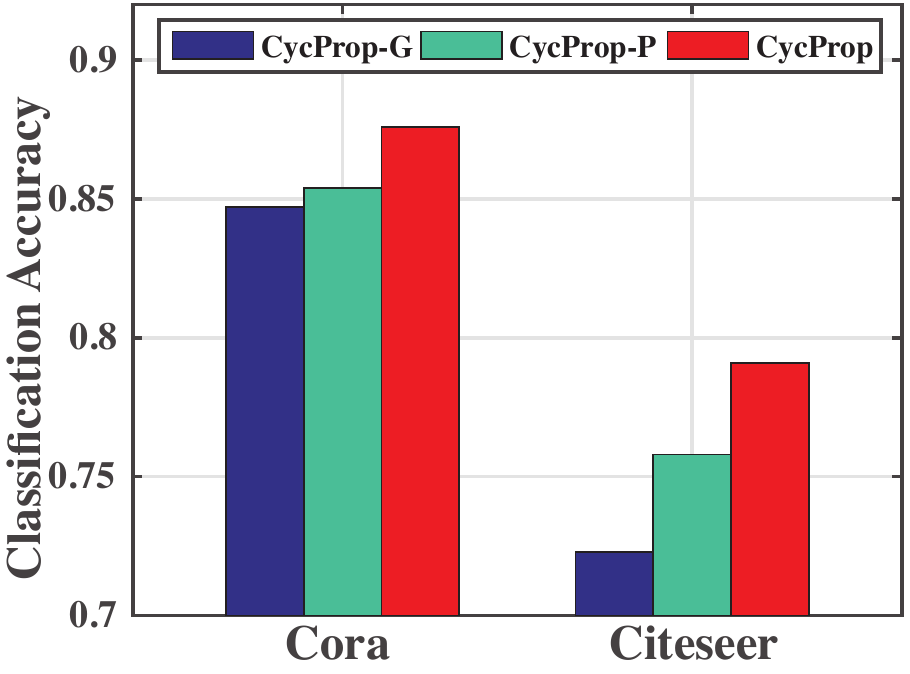}
	}
	\subfigure{
		\centering
		\includegraphics[width=0.45\textwidth]{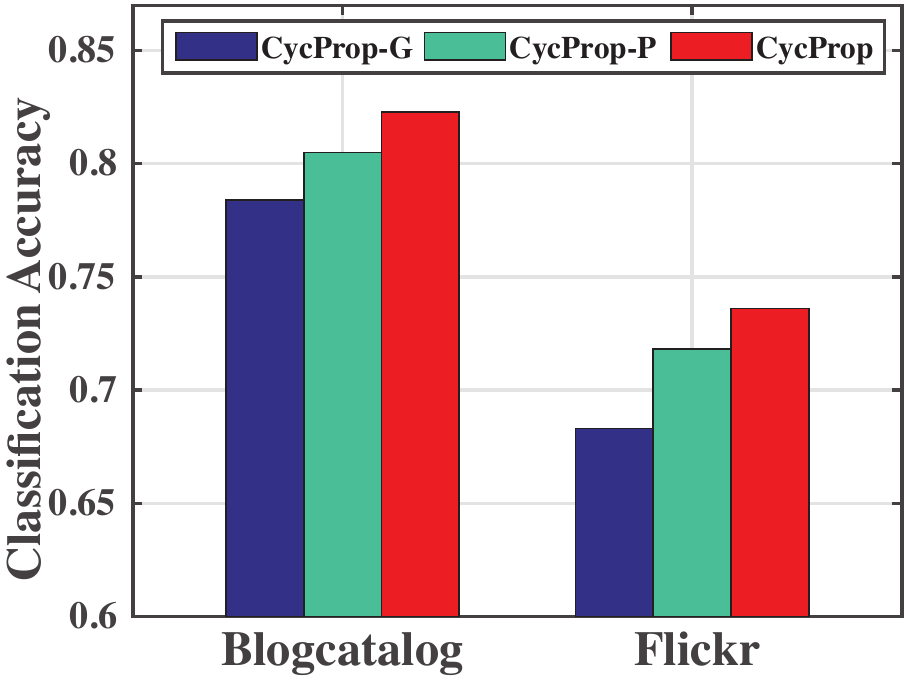}
	}
	\vspace{-0.1in}
	\caption{The performance of CycProp and its variants.}
	\label{fig:variants}
	\vspace{-0.2in}
\end{figure}

In this subsection, we investigate how each of the two individual components contributes to the performance of CycProp. Figure \ref{fig:variants} shows the classification results of CycProp and its two variants. Among them, CycProp-P denotes the case where only the label propagation module is used and CycProp-G indicates the case where only the graph neural network module is used. The best results are achieved by the full CycProp, which validates the effectiveness of combining these two modules in a mutually reinforcing manner. Moreover, CycProp-P always outperforms CycProp-G, thus the propagation module seems to play a more important role in the joint framework. This result verifies its advantage of propagating labels on graph rather than classifying each node independently. 

\subsection{Parameter Sensitivity}
\begin{figure}[!htb]\small
	\vspace{-0.1in}
	\centering
	\subfigure{
		\centering
		\includegraphics[width=0.45\textwidth]{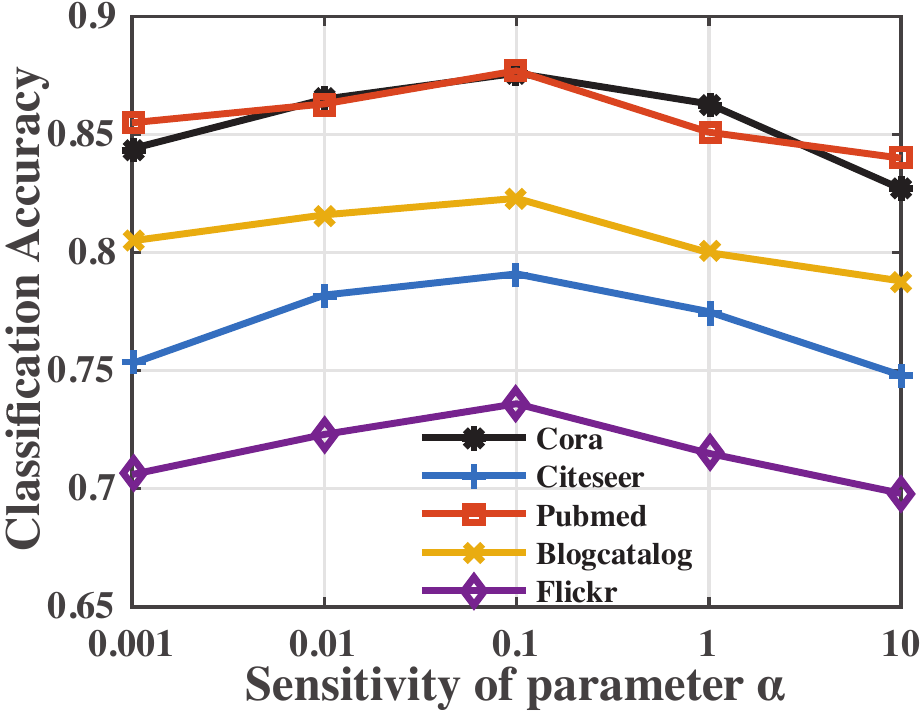}
	}
	\vspace{-0.1in}
	\subfigure{
		\centering
		\includegraphics[width=0.45\textwidth]{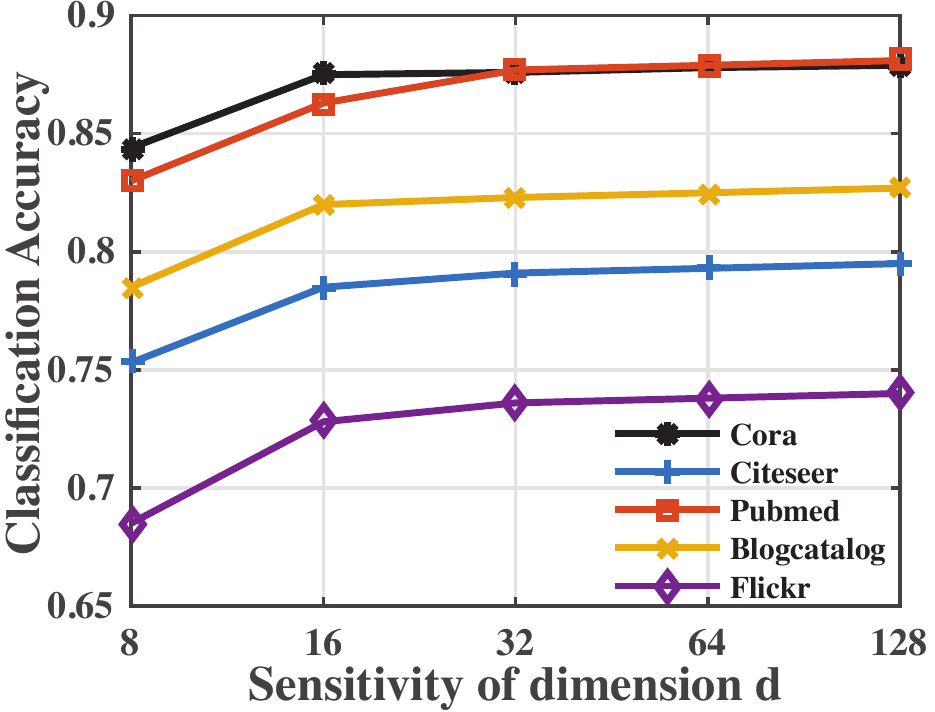}
	}
	\caption{The sensitivity of the trade-off parameter $\alpha$ and the node embedding dimension $d$.}
	\label{fig:parameters}
	\vspace{-0.2in}
\end{figure}

In this subsection, we study the impact of several parameters by varying them in different scales. Due to the limited space, we only show sensitivities of the trade-off parameter $\alpha$ of Equation (\ref{objective_function}) and the node embedding dimension $d$ in Figure \ref{fig:parameters}. As we can see,  $\alpha = 0.1$ is mostly the best across different data sets. When $\alpha$ is too small or too large, the performance becomes worse. For node embedding dimension $d$, we observe that, by increasing $d$, the performance first increases and then keeps stable. Besides, $\lambda$ is initialized with $0.1$ and $\mu = 10$, $\delta = 0.1$ usually get the best results and we don't observe too much difference when varying them.

\section{Conclusions}\label{sec:conclusion}
In this paper, we investigate the semi-supervised learning task on graphs and introduce a unified framework CycProp, which integrates label propagation and GNN in a cyclic and dynamically reinforcing manner. Specifically, in each iteration, we employ the graph neural network module to learn the informative node embeddings that can refine the edge weights for facilitating the label propagation; then the learned highly reliable labels obtained from the label propagation module are incorporated into the model to fine-tune the node embedding procedure, thus forming a closed cyclic training loop. Extensive experiments on five real-world datasets demonstrate the effectiveness of CycProp and its superiority to a range of state-of-the-art methods. In the future, we plan to extend our framework into heterogeneous information networks, where nodes and links are of different types.

\bibliographystyle{spmpsci}      
\bibliography{reference}

\end{document}